# A DIFFERENTIATED REWARD METHOD FOR REINFORCEMENT LEARNING BASED MULTI-VEHICLE COOPERATIVE DECISION-MAKING ALGORITHMS

[1]Ye, Han; [1]Lijun, Zhang*; [1]Dejian, Meng; [1]Zhuang, Zhang
[1]School of Automotive Studies, Tongji University, China

ABSTRACT
Deep reinforcement learning (DRL) shows great potential for optimizing multi-vehicle cooperative driving strategies through the state-action-reward feedback loop, but it still faces challenges such as low sample efficiency. This paper proposed a differentiated reward method based on steady-state transition systems, which incorporates state transition gradient into the reward function by analysing traffic flow characteristics, aiming to optimize action selection and boost policy learning in multi-vehicle cooperative decision-making. The performance of the proposed method is validated in RL algorithms such as MAPPO, MADQN, and QMIX under varying connected and automated vehicle (CAV) penetration rate. Results show that the differentiated reward method significantly accelerates training convergence and outperforms centering reward and other common reward shaping method in terms of traffic efficiency, safety, and action rationality. Additionally, the method demonstrates good adaptability to varying penetration rates of autonomous vehicles, providing a novel approach for multi-agent cooperative decision-making in mixed traffic scenarios. Code is available at https://github.com/leoPub/diff_rew.

## I. INTRODUCTION

With continuous advancements in perception technologies and local path planning for single-vehicle autonomous driving systems, coupled with rapid development of V2X communication and computing platforms, collaborative decision-making of connected and automated vehicles (CAVs) has demonstrated significant potential as a key approach to enhance traffic efficiency and road safety. Research indicates that in typical scenarios such as unsignalized intersections and highway merging zones, single-vehicle decision-making systems may lead to traffic efficiency degradation and safety risks due to insufficient global coordination capabilities [1]. The multi-vehicle cooperative decision-making systems hold crucial research value for constructing next-generation intelligent transportation system.

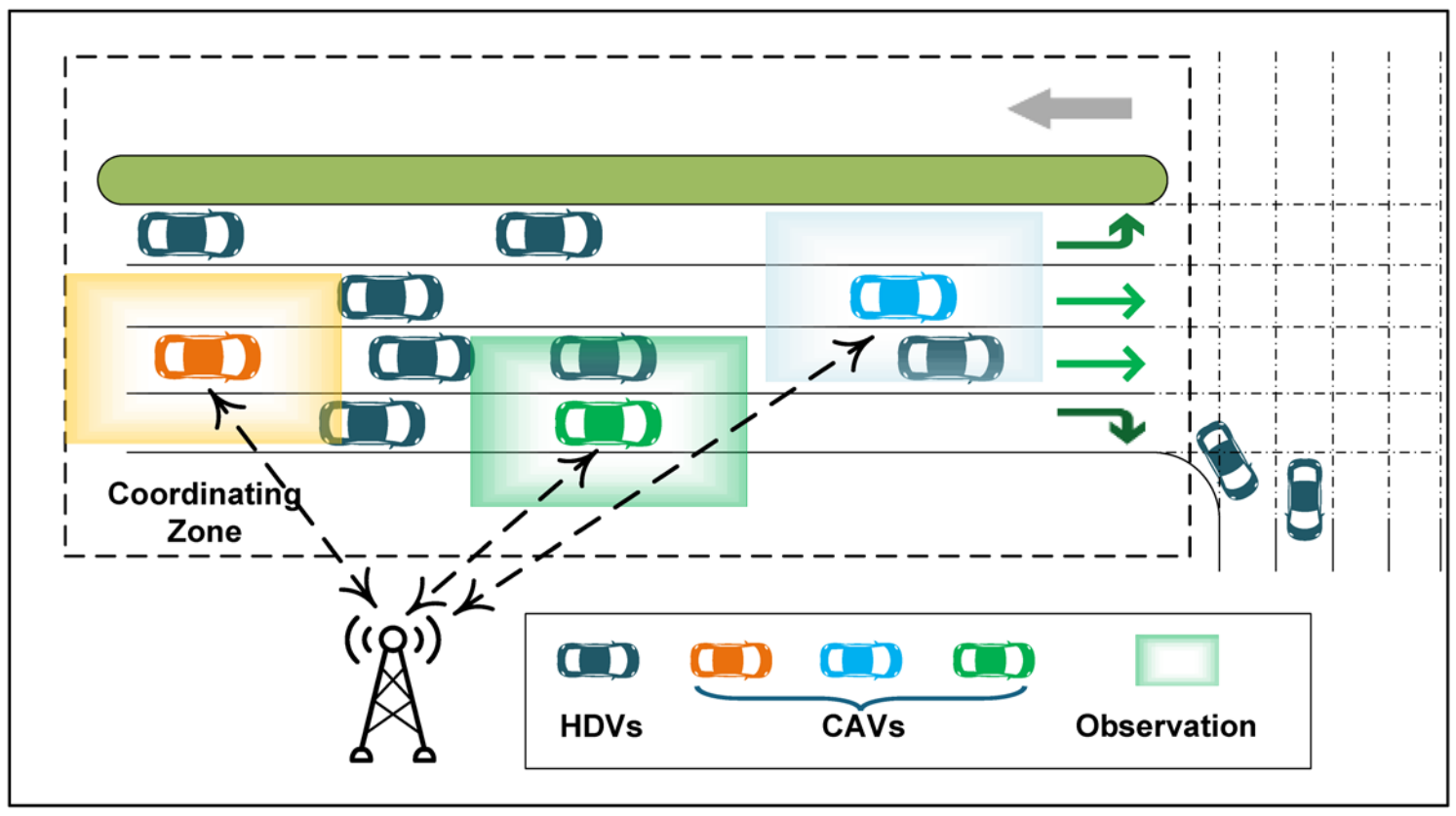

Figure 1: A schematic diagram illustrating multi-vehicle cooperative control in a mixed traffic environment with CAVs and human-driven vehicles (HDVs) under V2X communication. The framework highlights vehicle-to-vehicle (V2V) and vehicle-to-infrastructure (V2I) interactions for enhanced coordination and safety.

Deep reinforcement learning (DRL), with its self-adaptive learning capabilities in dynamic environments, has emerged as one of the mainstream methods for vehicle decision-making [2]-[4]. DRL-based vehicle decision-making algorithms have demonstrated significant improvements in critical metrics such as trajectory prediction accuracy and risk avoidance [2]. However, the application of Multi-Agent Reinforcement Learning (MARL) in multi-vehicle decision making (MVDM) still faces persistent challenges including low sample efficiency, the curse of dimensionality, and long-tail problems [5].

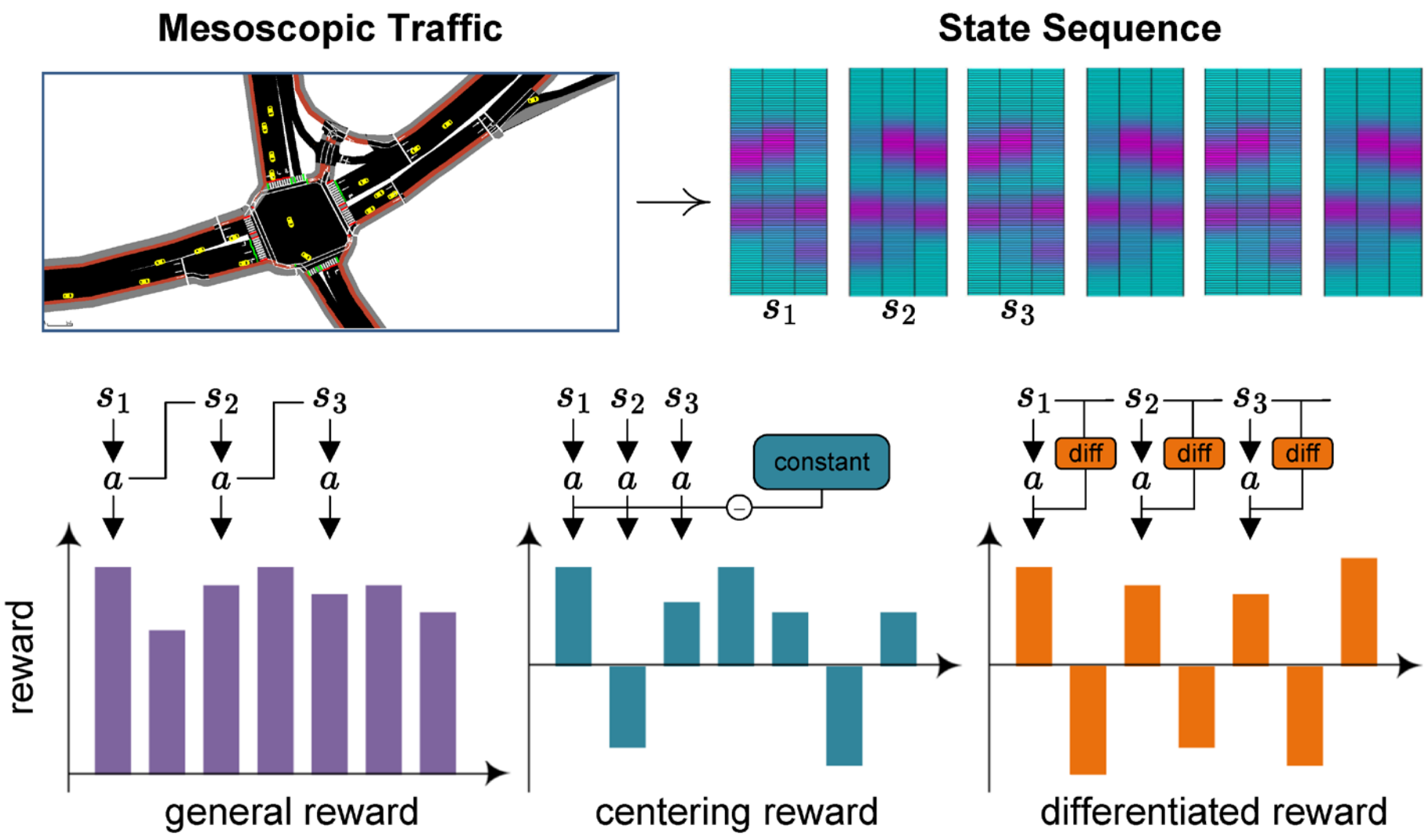

Figure 2: A comparative illustration of the general reward function (GR), centering reward function (CR), and differentiated reward function (DR). In most reinforcement learning studies, the reward function is explicitly expressed based on the new state entered after performing an action. The centralized reward function, on the base of the former, subtracts a baseline value from the reward. In contrast, the differentiated reward function proposed in this paper derives rewards by comparing the change in states between the current and previous time steps, in conjunction with the actions executed by the agent or system.

In MVDM researches, reward functions are typically formulated based on vehicle positions, velocities, and inter-vehicle interaction events (e.g., car-following, lane-changing negotiation, collision avoidance). Reward functions guide vehicles to make rational decisions, rendering its great importance in MVDM algorithms [6], [7]. Specifically, from a mesoscopic traffic flow perspective, vehicle states remain stable and evolve gradually over time in most scenarios. Under typical decision-making frequencies ($\geqslant$ 10 Hz), reinforcement learning algorithms may fail to distinguish between different actions in adjacent states due to data noises. This paper proposes a differentiated reward approach based on a steady-state transition system. Experimental results demonstrate that the proposed method significantly accelerated the training convergence speed of reinforcement learning and exhibited more rational behavior in action selection.

The main contributions of this paper can be summarized as follows:

1) A differentiated reward method in vehicle decision-making with steady-state transition is formulated from perspective of reinforcement learning theory. By employing differentiated reward, the performance, including overall performance and sample efficiency, of reinforcement learning algorithms in continuous multi-vehicle cooperative decision-making tasks is enhanced.
2) Thorough simulation experiments for multi-vehicle cooperative decision-making under different autonomous vehicle penetration rates in a continuous traffic flow environment is conducted, validating the scalability and learning stability of differentiated reward method in multi-agent traffic scenarios.

## II. RELATED WORKS

**DRL-based Multi-Vehicle Cooperative Decision-Making**: In recent years, researchers have devoted significant attention to MVDM problems in dynamic traffics. Traditional rule-based or classical control theory-based methods often exhibit limited scalability and adaptability in complex traffic scenarios. Increasingly mature multi-agent reinforcement learning (MARL) algorithms, such as MADQN, MADDPG, MAPPO, and QMIX, enable agents to learn optimal strategies through environmental interactions [4]. These algorithms have substantially enhanced the quality of multi-vehicle coordination in typical scenarios, including signal-free intersections [8]-[11], highway ramp merging [12]-[14], and mixed traffic scenarios [15], [16]. Zhuang H et al. proposed a distributed multi-agent proximal policy optimization (MAPPO) method with an attention-enhanced mechanism (Attn-MAPPO) for joint collision avoidance and efficient intersection traversal, demonstrating superior performance over heuristic rule-based models [8]. Chen D et al. formulated the mixed-traffic highway ramp merging problem as a MARL task, employing curriculum learning to effectively train agents for challenging driving maneuvers [13]. Current research demonstrates that RL-based multi-vehicle cooperative decision-making algorithms have achieved performance comparable to, or even exceeding, that of human drivers in simulated environments or specific benchmarks. However, enhancing training sample efficiency and stability remains an open problem.

**Reward Function Formulation in DRL**: In reinforcement learning, the reward function plays a pivotal role in guiding the learning process of agents. A properly formulated reward function can significantly enhance algorithm performance [17], [18], particularly in complex multi-vehicle collaborative decision-making problems, where incorporating problem-specific characteristics into reward design is critical.

Reward shaping is a commonly employed technique in reinforcement learning (RL) research that incorporates domain knowledge into RL frameworks to accelerate the discovery of optimal policies [19]. R. D. et al. [20] proposed a straightforward framework for integrating shaped rewards into RL systems and provided analytical tools to demonstrate that specific reward shaping choices can significantly enhance sample efficiency. Abhishek N et al. [21] introduces a generally applicable reward centering approach, which substantially improves the performance of RL algorithms under standard discount factors. However, this approach struggles to handle scenarios where reward functions contain substantial underlying offsets.

Based on the aforementioned research, this study proposes a differentiated reward mechanism to enhance multi-vehicle cooperative decision-making capabilities in RL. We establish a theoretical framework for differentiated reward reinforcement learning and demonstrate its efficacy through concrete case studies. Specifically, in continuous traffic flow environments, the proposed method optimizes inter-agent reward allocation during steady-state transitions, thereby enhancing sample efficiency, accelerating convergence and enhancing training stability.

## III. PROBLEM FORMULATION

### A. Markov Decision Process

We model the interaction between the agent and the environment using a finite Markov Decision Process (MDP) $(\mathcal{S},\mathcal{A},\mathcal{R},p)$, where $\mathcal{S}$ , $\mathcal{A}$ , and $\mathcal{R}$ represent the state space, action space, and reward space, respectively. The state transition probability is defined as $p(s',r| s,a)=\Pr(S_{t+1}=s',R_{t+1}=r| S_t=s,A_t=a)$ , with the map $p:\mathcal{S}\times\mathcal{R}\times\mathcal{S}\times\mathcal{A}\rightarrow[0,1]$. At each time step $t$ , the agent is in state $S_t\in\mathcal{S}$ and selects an action $A_t\in\mathcal{A}$ using a behavior policy $b:\mathcal{A}\times\mathcal{S}\rightarrow[0,1]$. Based on this transition, the system moves to the next state $S_{t+1}\in\mathcal{S}$ , and the agent receives a reward $R_{t+1}\in\mathcal{R}$ . In the continuous problem we consider, the interaction between the agent and the environment persists indefinitely. The agent's goal is to maximize the long-term average reward. To achieve this, we estimate the expected discounted sum of rewards for each state under a target policy $\pi$ , which is $v_{\pi}^{\gamma}(s)\doteq E[\sum_{t=0}^{\infty}\gamma^{t}R_{t+1}| S_t=s,A_{t:\infty}\sim\pi],\forall s$ , where $\gamma\in[0,1)$ .

### B. World Model

This paper addresses the multi-vehicle cooperative decision-making problem in urban traffic scenarios characterized by continuous traffic flow and mixed autonomy.

We consider a unidirectional branch of a bidirectional 8-lane road, where vehicles in all 4 lanes are randomly assigned one of three objectives: going straight, turning left, or turning right. The models for human-driven vehicles (HDVs) in terms of going straight and lane-changing follow the same settings as in our previous work [22].

For connected and autonomous vehicles (CAVs), as illustrated in Figure 1, each vehicle has accurate perception of its own position, speed, target lane, and vehicle type, as well as those of vehicles within its observation range. Additionally, CAVs can share their perception information through infrastructure. This means that CAVs within the coordination zone can access and utilize the perception information of all CAVs in the environment.

### C. Observation Space

For vehicle $i$ , its own state vector is defined as:

$$s_i^{\text{self}}=[p_i^{\text{lon}},p_i^{\text{lat}},v_i,\tau_i,g_i,d_i^{\text{left}},d_i^{\text{front}},d_i^{\text{right}}], \tag{1}$$

where $p_i^{\text{lon}} \in \mathbb{R}$ and $p_i^{\text{lat}} \in \mathbb{R}$ represent the longitudinal and lateral positions of the vehicle in the global coordinate system, respectively, $v_i \in \mathbb{R}^+$ denotes the current driving speed, $\tau_i \in \mathbb{Z}^+$ is the discretized vehicle type encoding, and $g_i \in \{0,1\}^k$ represents the driving goal (going straight, turning left, or turning right) using one-hot encoding. $d_i^{\text{left}}, d_i^{\text{front}}, d_i^{\text{right}} \in \mathbb{R}^+$ denote the distances to the nearest vehicles in the left, front, and right lanes, respectively. If no corresponding vehicle exists, the distance is set to a predefined maximum value $d_{\max}$.

For the set of surrounding vehicles $\text{nbr}^i = \{j| \ \| p_i - p_j \|_2 \le R\}$ (where $R$ is the perception radius), we construct the following relative state matrix:

$$M_i^{\text{nbr}} = \underset{j \in \text{nbr}^i}{\text{concat}} \left[ \Delta p_{ij}^{\text{lon}}, \Delta p_{ij}^{\text{lat}}, \Delta v_{ij}, \Delta \tau_{ij}, \Delta g_{ij} \right], \tag{2}$$

where $\Delta p_{ij}^{\text{lon}} = p_j^{\text{lon}} - p_i^{\text{lon}}$ and $\Delta p_{ij}^{\text{lat}} = p_j^{\text{lat}} - p_i^{\text{lat}}$ represent the relative positional relationship, $\Delta v_{ij} = v_j - v_i$ denotes the velocity difference, $\Delta \tau_{ij} = \tau_j - \tau_i$ reflects the vehicle type difference, and $\Delta g_{ij} = \| g_j - g_i \|_2$ is the Euclidean distance representing the difference in driving goals. When the number of surrounding vehicles $n < N_{max}$, the missing rows are padded with zero vectors.

The observation space is finally represented as

$$o_i = \text{concat}\left[ s_i^{\text{self}}, M_i^{\text{nbr}} \right]. \tag{3}$$

D. Action Space

The longitudinal control action set is defined as

$$\mathcal{A}_i^{\text{lon}} = \{a^{\text{acc}}, a^{\text{keep}}, a^{\text{dec}}\}, \tag{4}$$

where $a^{\text{acc}} \in \mathbb{R}^+$ represents acceleration, $a^{\text{keep}}$ denotes maintaining the current speed, and $a^{\text{dec}} \in \mathbb{R}^-$ represents deceleration. The longitudinal action is converted into actual acceleration execution through a dynamics model, and the vehicle's speed at the next time step is given by

$$\dot{v}_i = \text{clip}(v_i + a^{\text{lon}} \cdot \Delta t, 0, v_{\max}), \tag{5}$$

where $\Delta t$ is the decision time step, and the $\text{clip}(\cdot)$ function ensures that the speed is constrained within the range $[0, v_{\max}]$.

The lateral control action set is defined as

$$\mathcal{A}_i^{\text{lat}} = \{a^{\text{left}}, a^{\text{hold}}, a^{\text{right}}\}, \tag{6}$$

where $a^{\text{left}}$, $a^{\text{hold}}$, and $a^{\text{right}}$ represent changing to the left lane, maintaining the current lane, and changing to the right lane, respectively. The execution of lateral actions satisfies the lane boundary constraints:

$$L_i^{t+1} = \begin{cases} \max(L_i^t - 1, 1) & \text{if } a^{\text{lat}} = a^{\text{left}} \\ L_i^t & \text{if } a^{\text{lat}} = a^{\text{hold}} \\ \min(L_i^t + 1, N_{lane}) & \text{if } a^{\text{lat}} = a^{\text{right}} \end{cases}, \tag{7}$$

where $L_i^t \in \mathbb{Z}^+$ denotes the lane number of the vehicle at time $t$, and $N_{\text{lane}}$ is the total number of lanes on the road. If the target lane does not exist, the action automatically defaults to $a^{\text{hold}}$.

The complete action of agent $i$ at the decision-making moment is the Cartesian product of the longitudinal and lateral actions:

$$\mathcal{A}_i = \mathcal{A}_i^{\text{lon}} \times \mathcal{A}_i^{\text{lat}} \tag{8}$$

The design of the reward function is detailed in Section IV-C.

IV. METHODOLOGY

A. Method of Reward Centering

First, we describe the general idea of reward centering [21]. Reward centering involves subtracting the empirical mean of the rewards from the observed rewards, thereby achieving a mean-centered effect for the modified rewards.

For general reinforcement learning algorithms, we can perform a Laurent decomposition on the value function: The discounted value function can be decomposed into two parts, one of which is a constant that does not depend on the state or action, and thus does not affect the selection of actions. For a policy $\pi$ corresponding to the discount factor $\gamma$, the tabular discounted value function $h_\pi^\gamma : \mathcal{S} \to \mathbb{R}$ can be expressed as

$$h_\pi^\gamma(s) = \frac{r(\pi)}{1-\gamma} + \tilde{h}_\pi(s) + e_\pi^\gamma(s), \tag{9}$$

where $r(\pi)$ is the state-independent average reward obtained by policy $\pi$, and $\tilde{h}_\pi(s)$ is the differential value of state $s$. For ergodic Markov decision processes, these two terms can be defined as Equation (10).

$$\begin{aligned} r(\pi) \quad &\doteq \lim_{n\to\infty} \frac{1}{n} \sum_{t=1}^{n} \mathbb{E}\left[R_t \mid S_0, A_{0:t-1} \sim \pi\right], \\ \tilde{h}_\pi(s) \quad &\doteq \mathbb{E}\left[\sum_{k=1}^{\infty} \left(R_{t+k} - r(\pi)\right) \mid S_t = s, A_{t:\infty} \sim \pi\right]. \end{aligned} \tag{10}$$

Additionally, $e_\pi^\gamma(s)$ in Equation (9) represents an error term that approaches zero as the discount factor approaches 1. To distinguish the speed $v$, we do not use the common value function notation in reinforcement learning research but instead use $h$ to denote the value function.

In many reinforcement learning problems, the state-independent offset can be quite large. Consider subtracting the constant offset from each state's discounted value, i.e., $h_\pi^\gamma(s) - r(\pi)/(1-\gamma)$, which is referred to as the centered discounted value. The centered discounted value is much smaller in magnitude and changes very little as the discount factor increases. For most reinforcement learning problems (especially long-term ones), when the discount factor approaches 1, the magnitude of the discounted value increases dramatically, while the centered discounted value changes less and approaches the state differential value. The relevant representation of the discounted value function is as follows:

$$\begin{aligned} \tilde{h}_\pi^\gamma(s) \quad &\doteq \mathbb{E}\left[\sum_{t=0}^{\infty} \gamma^t \left(R_{t+1} - r(\pi)\right) \mid S_t = s, A_{t:\infty} \sim \pi\right], \\ h_\pi^\gamma(s) \quad &= \frac{r(\pi)}{1-\gamma} + \overbrace{\tilde{h}_\pi(s) + e_\pi^\gamma(s)}^{\tilde{h}_\pi^\gamma(s)}, \end{aligned} \tag{11}$$

where $\gamma \in [0,1]$. When $\gamma = 1$, the centered discounted value and the differential value are identical, i.e., $\tilde{h}_\pi^\gamma(s) = \tilde{h}_\pi(s), \forall s$. More generally, the centered discounted value is the differential value plus the expansion error of the Laurent series, as shown in the second part of Equation (11).

Therefore, reward centering allows reinforcement learning algorithms to capture all information in the discounted value function through two components: (1). the constant average reward and (2). the centered discounted value function. The role of this decomposition is highly intuitive:

a) When $\gamma \to 1$, the discounted value tends to explode, but the centered discounted value remains small and manageable.
b) If the rewards of the reinforcement learning problem are shifted by a constant $c$, the magnitude of the discounted value will increase by $c/(1-\gamma)$, but the centered discounted value remains unchanged because the average reward increases by $c$.

B. Differentiated Reward Formulation and Analysis

Define a Markov chain $\mathcal{S} = \{S(t) : t = 0,1,2,\ldots\}$, whose state space is $\Re^\ell$, and $\mathcal{S}$ evolves according to a nonlinear state-space model:

$$S(t+1) = \mathcal{T}(S(t), N(t+1)), \quad t \geq 0, \tag{12}$$

where $N$ is an m-dimensional disturbance sequence modeling the nonlinear state transition, composed of independent and identically distributed random variables, $\mathcal{T} : \Re^{\ell+m} \to \Re^\ell$ is a continuous mapping. Under this assumption, for all $t \geq 0$, $S(t+1)$ is a continuous function of the initial condition $S(0) = s_0$.

Under the discount factor $\gamma \in (0,1)$, the discounted value function can be written as:

$$h_{\pi}^{\gamma}(s_0) := \sum_{t=0}^{\infty} \gamma^t \mathbb{E}[R(t) | R(0) = r_0], \quad r_0 \in \Re^{\ell}. \tag{13}$$

The goal of TD learning is to approximate $h^{\gamma}$ as an element of a function family $\{h_{\pi(\theta)}^{\gamma} : \theta \in \Re^d\}$. Here, we assume that the discounted value function can be linearized in the following form:

$$h_{\pi(\theta)}^{\gamma} = \sum_{j=1}^{d} \theta_j \psi_j, \tag{14}$$

where $\theta = (\theta_1, \theta_2, \ldots, \theta_d)^T$, $\psi = (\psi_1, \psi_2, \ldots, \psi_d)^T$, and the given set of basis functions $\psi : \Re^{\ell} \to \Re^d$ is assumed to be continuously differentiable. Indeed, most reward function expressions derived from the analytical relationships of vehicle dynamics satisfy this condition.

The goal of TD learning can be expressed as a minimum norm problem:

$$\theta^* = \underset{\theta}{\operatorname{argmin}} \| h_{\pi(\theta)}^{\gamma} - h^{\gamma} \|_{\pi}^2. \tag{15}$$

Assume that the value function $h^{\gamma}$ and all its possible approximations $\{h_{\pi(\theta)}^{\gamma} : \theta \in \Re^d\}$ are continuously differentiable as functions of the state $s$, i.e., for each $\theta \in \Re^d$, $h^{\gamma}, h_{\pi(\theta)}^{\gamma} \in C^1$. Based on the linear parameterization in Equation (14), we obtain the following form of the differential value function:

$$\nabla h_{\pi(\theta)}^{\gamma} = \sum_{j=1}^{d} \theta_j \nabla \psi_j, \tag{16}$$

where the gradient is taken with respect to the state $s$.

Now, the goal of TD learning changes accordingly to:

$$\theta^* = \underset{\theta}{\operatorname{argmin}} \| \nabla h_{\pi(\theta)}^{\gamma} - \nabla h^{\gamma} \|_{\pi}^2. \tag{17}$$

Evidently, in conventional reinforcement learning algorithms, directly utilizing the derivative of the reward function is essentially equivalent to solving the problem posed by Equation (17).

It can be observed that both reward centering and the reward differentiating method exhibit translation invariance to shifts in the reward function (removing the mean or computing gradients can eliminate the effects of constant offsets). When $\gamma \to 1$, the long-term fluctuations of $h_{\pi(\theta)}^{\gamma}(s)$ under steady-state conditions in reward centering are averaged out, causing the value function to be predominantly governed by $r(\pi)/(1-\gamma)$. The centered value function satisfies Bellman equation $\tilde{h}_{\pi}^{\gamma}(s) = \mathbb{E}\left[R - r(\pi) + \gamma \tilde{h}_{\pi}^{\gamma}(s') | s\right]$, while the gradient objective $\nabla h_{\pi}^{\gamma}(s)$ in the reward differentiating method aligns with the solution of the steady-state Bellman equation.

Furthermore, the reward differentiating method exhibits superior performance over reward centering when addressing scenarios involving global constant offsets or state-dependent systematic biases in the reward function. Consider a reward function with a global constant offset $c$. Reward centering requires precise estimation of $r(\pi) = c + \mathbb{E}[R_t]$, where significant estimation errors under large offsets may introduce biases in value function estimation. For state-dependent systematic biases (e.g., $R(s) = R_{\text{true}}(s) + c(s)$, where $c(s)$ denotes a state-dependent offset function), reward centering necessitates estimating the average offset $\mathbb{E}[c(s)]$ for each state, which becomes computationally intractable in complex environments. In contrast, the gradient $\nabla h^{\gamma}(s)$ is driven by difference of rewards between adjacent states. Even when state-dependent offsets $c(s)$ exist, as long as the offset function varies smoothly across local state transitions (e.g., $c(s') - c(s)$ remains small), the gradient updates can still approximate the true reward difference $R_{\text{true}}(s') - R_{\text{true}}(s)$.

C. Differentiated Reward Implementation in MVDM

In MVDM, the reward function commonly used can be expressed as Equation (18) [24]-[26].

$$
\begin{aligned}
R &= w_1 R_{\text{speed}} + w_2 R_{\text{intention}} + w_3 P_{\text{collision}} + w_4 P_{LC} \\
&= \frac{1}{N}\left( w_1 \sum_{i=1}^{N} \frac{v_i}{v_{\max}} + w_2 N_{\text{sat}} + w_3 N_{\text{col}} + w_4 N_{LC} \right),
\end{aligned} \tag{18}
$$

where $N$ is the number of vehicles in the scene (including HDVs and CAVs), $N_{\text{onramp}}$ is the number of vehicles passing through the intention area at the previous time step and aiming for the ramp, $N_{\text{collision}}$ is the number of collisions, and $N_{LC}$ is the number of frequently lane-changing vehicles.

Based on the proposed differentiated reward method, we have improved the reward function in Equation (18) as follows. Due to the variable number of agents in the setup, this paper cannot directly use the sum of the local rewards of all agents as the reward function. Instead, an index needs to be designed to objectively evaluate the overall traffic quality within the coordinating zone. To this end, we design the following reward function:

$$
r_{\text{env}} = \frac{1}{|\mathcal{N}_{\text{CAV}}|} \sum_i \left( \omega_1 r_a^i + \omega_2 r_p^i \right) + \omega_3 r_{\text{flow}} + \omega_4 r_{\text{safe}}, \tag{19}
$$

where $r_a^i$ and $r_p^i$ represent the action reward and position reward for CAV $i$, respectively, $r_{\text{flow}}$ is used to evaluate the overall traffic flow speed, and $r_{\text{safe}}$ is the traffic safety indicator. The parameters $\omega_{1,2,3,4}$ are the weights assigned to each reward component. Specifically:

$$
r_a^i = \begin{cases} 1 & \text{if vehicle } i \text{ accelerating or keeping highspeed,} \\ 0 & \text{otherwise.} \end{cases} \tag{20}
$$

We design a potential field $f_p^i(x, y)$ based on vehicle $i$'s longitudinal position and the lane it occupies, to evaluate the value of the current position of it relative to its target.

$$
f_p^i(x, y) = \frac{e^{-\frac{(l-x)^2}{2\sigma^2}}}{\zeta \left| y_{tar}^i - y \right| + 1}, \tag{21}
$$

where $\sigma$ and $\zeta$ are the longitudinal and lateral decay coefficients, respectively. Based on the concept of reward differentiating, we define the position reward as:

$$
r_p^i = \mathbf{v}^i \cdot \nabla f_p^i(x, y), \tag{22}
$$

where $\mathbf{v}^i = [v_x^i, v_y^i]$. In this paper, $v_y^i \in \{-1, 0, 1\}$, so we have the discrete form of Equation (22):

$$
r_p^i = \left[ v_x^i (l - x) + \frac{\zeta v_y^i \operatorname{sign}\left( y - y_{tar}^i \right)}{\zeta \left| y_{tar}^i - y \right| + 1} \right] \cdot f_p^i(x, y), \tag{23}
$$

here, we define that when $y^i = y_{\text{tar}}^i$, $\operatorname{sign}\left( y - y_{\text{tar}}^i \right) = -v_y^i$. It can be observed that the terms $\omega_1 r_a^i$ and $\omega_2 r_p^i$ sometimes have the same reward effect. In this case, we treat them as strengthen of the action reward without making a more detailed distinction.

Although the potential function in this paper is designed for multi-lane highway scenarios, its core idea—constructing rewards based on the positional relationship between vehicles and dynamic targets—has potential for extension. For instance, at unsignalized intersections, the 'target lane' $y_{\text{tar}}^i$ could be replaced by a dynamically planned sequence of waypoints through the intersection. This provides a theoretical foundation for handling more complex topological structures.

We use the overall speed of the traffic flow to evaluate the current traffic volume, which is:

$$
r_{\text{flow}} = \frac{1}{|\mathcal{N}|} \sum_i \frac{v^i}{v_{max}}. \tag{24}
$$

For the safety indicator, since accidents are rare events and their occurrence typically has a significant impact on overall traffic, we use summation rather than averaging, that is

$$
r_{\text{safe}} = \sum_i \mathbb{I}(i), \tag{25}
$$

where $\mathbb{I}(i)$ is the indicator function, which equals 1 if vehicle $i$ is involved in a collision, and 0 otherwise.

## V. EXPERIMENT

### A. Simulation Environment and Experiment Settings

The experiments are based on the open-source microscopic traffic simulation platform SUMO (Simulation of Urban Mobility), which supports high-precision vehicle dynamics modeling, multi-agent collaborative control, and visual analysis of complex traffic scenarios [27]. The simulation scenario is a unidirectional branch of a bidirectional 8-lane highway (4 lanes in the same direction), with a straight road segment length of 250 meters, and a speed limit of 25 m/s (90 km/h). Vehicular traffic generation is achieved through SUMO's native traffic flow module, which employs a Poisson-distributed stochastic process to inject background vehicles. The baseline traffic density is maintained at 250 vehicles/(h·lane), aligning with empirically observed traffic patterns on urban arterials during off-peak periods. Once inserted in traffic, each vehicle is randomly assigned one of 3 objectives: going straight, turning left, or turning right at the end of the road. Accordingly, the target lane is the middle 2 lanes, the left most lane, and the right most lane, respectively.

Table 1: Key simulation parameters

| Parameter Category | Parameter Value |
|---|---|
| Road Length $l$ | 250 m |
| Number of Lanes | 4 (unidirectional) |
| Traffic Density | 250 / (h·lane) |
| Maximum Vehicle Speed | 25 m/s |
| $d_{\max}$ in Equation (1) | 1000 |
| Longitudinal Observation Range | $\pm$100 m |
| Lateral Observation Range | $\pm$1 adjacent lanes |
| Vehicle Objective | 1/3 each for 3 directions |
| Episode Duration | 18 s |
| Decision Frequency | 10 Hz |
| Autonomous Vehicle Penetration Rate | 25%, 50%, 75%, 100% |

### B. Compared Methods

We compare the training and deployment performance of widely used multi-agent reinforcement learning algorithms, including MADQN [24], MAPPO [28], and QMIX [29], by employing the generally adopted vehicle decision reward function (GR, with Equation (18)), the centering reward function (CR, with oracle centering in [21]), and the differentiated reward function (DR, with Equation (23)).

All algorithms share the following hyperparameter settings in Table 2.

Table 2: Algorithm hyperparameter settings

| Hyperparameter | Value |
|---|---|
| Total episodes | 50000 |
| Episode Duration | 18 s (180 steps) |
| RMSProp Learning Rate | $3\times10^{-4}$ |
| RMSProp $\alpha$ | 0.99 |
| RMSProp $\varepsilon$ | $1\times10^{-5}$ |
| Batch Size | 32 (episodic) |
| Discount Factor $\gamma$ | 0.98 |
| Replay Buffer Size | $1\times10^{5}$ |
| Target Network Update Interval | 10 episodes |
| Exploration Rate Decay | 0.998 (linear, episodic) |
| Weights $\omega_{1,2,3,4}$ in Equation (19) | 10, 1e3, 1, -5 |
| Reward for Arrival | 30 |
| $\sigma$, $\zeta$ in Equation (21) | 60, 1 |

As discussed here and Section V-A, under a traffic flow density of 250 vehicles/(h·lane), this study conducted training experiments using 3 reward functions (GR, CR, DR) for MADQN, MAPPO, and QMIX algorithms under 4 distinct CAV penetration rate conditions: 25%, 50%, 75%, and 100%. For each algorithm in every scenario, three independent training trials were executed with three distinct random seeds. The training process is comprehensively presented in Figure 3.

C. Evaluation Metrics

**Average Speed (Avg. Speed)**, reflecting overall traffic efficiency:

$$\overline{v} = \frac{\sum_{t=0}^{\text{END}} \overline{v}_t}{\sum_{t=0}^{\text{END}} 1}, \tag{26}$$

where $\overline{v}_t$ is the average speed of all vehicles on road at time $t$.

**Minimum Headway (Min. Gap)**, quantifying potential collision risk:

$$C_{\text{GAP}} = \frac{\sum_{e=0}^{N_{\text{epi}}} g_{\text{min},\, e}}{N_{\text{epi}}}, \tag{27}$$

where $g_{\text{min},\, e}$ is the minimum gap of vehicle in test episode $e$, $N_{\text{epi}}$ is the total number of test episodes.

**Lane Change Frequency (LC. Freq.)**, evaluating the rationality of lane resource utilization:

$$F_{\text{LC}} = \frac{\sum_{t=0}^{\text{END}} n_{\text{LC},t}}{\sum_{t=0}^{\text{END}} n_{\text{CAV},t}}, \tag{28}$$

where $n_{\text{LC},t}$ is the number of lane changes action of simulation step $t$, and $n_{\text{CAV},t}$ is the CAV number of time $t$. It is crucial to note that frequent lane-changing behavior are irrational and should be avoided from the perspectives of safety, comfort, and vehicle energy utilization efficiency. Consequently, under the premise of satisfying basic driving safety and speed requirements, a smaller $F_{\text{LC}}$ indicates a more rational action selection capability of the vehicle.

**Mission Success Rate (Succ. Rate)**:

$$\textbf{SR}. = \frac{\text{number of vehicles reach their target}}{\text{number of CAVs}}. \tag{29}$$

D. Result and Comparison

Figure 3 illustrates the training process of the algorithms. To better capture the early-stage convergence and later-stage stability, a logarithmic scale was applied to the horizontal axis. During the training process, the bias correction for CR was implemented when recording return values to enable direct comparison with GR's training progression. For DR, due to the inability to align return values with GR and CR under a unified metric through simple transformations, dual y-axes were employed for data logging. Consequently, GR and CR share the right y-axis, facilitating comparisons of absolute return values and training trend. The training curves of all 3 algorithms are plotted in a single figure to primarily compare convergence speed and stability. It is noteworthy that when presenting the final graphical results, we aligned the left and right axes based on the quality of the converged policies. Specifically, if a converged curve appears higher, we ensured that statistically significant return test results would be larger within the displayed range. Due to the nonlinear numerical mapping between differential rewards and the other two reward methods, relative relationships may exhibit inconsistencies across the entire vertical coordinate range. Quantitative performance evaluations of the algorithms are conducted during the model testing phase.

It can be observed in Figure 3 that in MADQN, the training curves for all three reward functions exhibit significant fluctuations, indicating poor stability. Despite tuning the algorithm to its optimal state, the performance of the three reward functions varies across different penetration rates. This instability is likely due to MADQN being based on independent Q-learning, which struggles to adapt to multi-vehicle cooperative decision-making tasks. We will discuss this phenomenon in detail in the later sections. Additionally, value-iteration algorithm like DQN have more reward variance than another rewards, which may also lead to the fluctuations. The high variance of value iteration methods also manifests in the QMIX implementation. In MAPPO, although the algorithm achieves stable convergence, the performance of the centralized reward function exhibits considerable deviation in the stabilized phase under the same settings. Specifically, for penetration rates of 0.25, 0.5, and 1.0, the orange curve shows substantial deviation after 20,000 episodes.

It is evident that the convergence speed of the differentiated reward function is significantly faster than that of the centralized and general reward functions, especially in experiments with MAPPO and QMIX. Additionally, the return values of the centralized reward function are noticeably higher than those of the general reward function, particularly in the later training stages (> 20,000 episodes). This finding is consistent with the conclusions reported in [21].

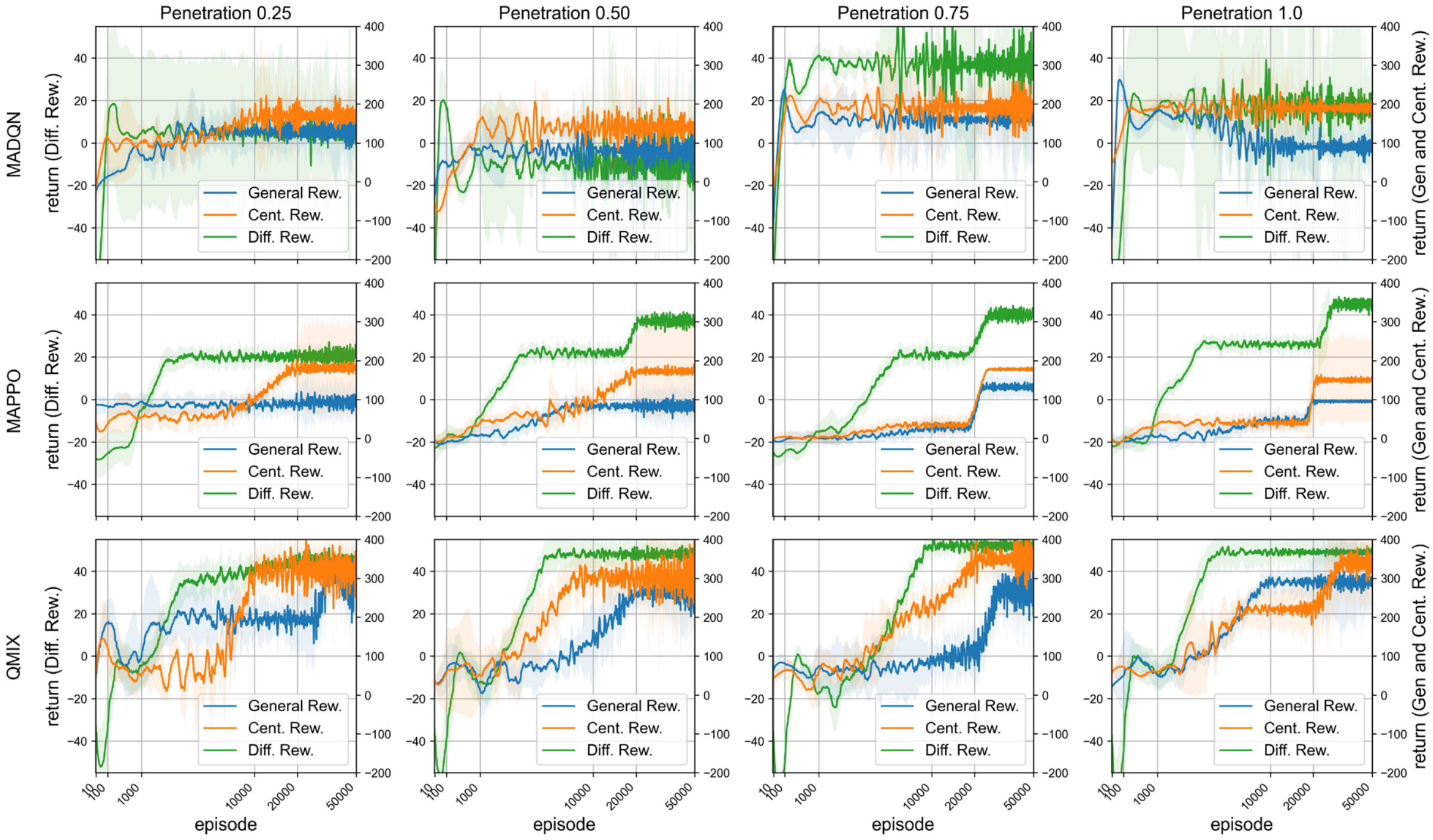

Figure 3: Training comparison. To better illustrate the early-stage convergence and later-stage stability of the algorithms, a logarithmic scale was used for the horizontal axis. Algorithm performance was evaluated under 3 reinforcement learning algorithms (MADQN, MAPPO, and QMIX) and 4 penetration rates (0.25, 0.50, 0.75, and 1.0) using 3 types of reward functions: differentiated reward function (green curve), centralized reward function (orange curve), and general reward function (blue curve). In each subplot, the green curve corresponds to the left vertical axis, while the other two curves correspond to the right vertical axis. The left and right axes are aligned based on the quality of the converged policies.

Table 3: Metrics comparison of reward function in QMIX

| Pene. Rate | Rew. Fn. | Avg. Speed (m/s) | Min. Gap (m) | LC. Freq. (time/min) | Succ. Rate (%) |
|---|---|---|---|---|---|
| 25% | GR | 7.95±0.42 | 1.33±0.74 | 0.67±0.03 | 49.20±0.24 |
| | CR | 10.95±0.33 | 5.36±0.62 | 0.14±0.02 | 61.45±0.15 |
| | DR | **11.46±0.18** | **36.47±0.36** | **0.07±0.02** | **96.49±0.12** |
| 50% | GR | 7.19±0.83 | 4.26±0.55 | 0.14±0.01 | 33.72±2.66 |
| | CR | **14.18±0.41** | 5.55±0.91 | 0.10±0.04 | 68.83±0.19 |
| | DR | 12.06±0.28 | **49.04±0.68** | **0.04±0.03** | **97.01±0.14** |
| 75% | GR | 6.31±1.37 | 6.01±4.51 | 0.12±0.04 | 42.16±5.60 |
| | CR | 10.60±2.44 | 5.22±1.87 | 0.11±0.45 | 73.78±9.18 |
| | DR | **12.01±0.35** | **53.56±0.74** | **0.01±0.00** | **94.64±0.08** |
| 100% | GR | 11.89±1.07 | 6.76±1.87 | **0.21±0.08** | 34.37±2.74 |
| | CR | 12.58±0.42 | 5.24±0.86 | 0.15±0.04 | 62.88±0.23 |
| | DR | **16.79±0.33** | **32.30±0.72** | 0.10±0.03 | **88.82±0.07** |

Table 3 shows the performance metrics of the 3 reward functions under the QMIX algorithm at different penetration rates. For each algorithm under every simulation configuration, we conducted three independent trials, with each trial comprising 1000 episodes and utilizing the optimal model obtained in the training phase.

It can be observed that in most cases the differentiated reward function achieves the best performance in terms of traffic efficiency, safety, action rationality, and task completion rate.

The DR maintains relatively high speeds (11.46-16.79 m/s) and safe spacing (32.30-53.56 m) across Pene. Rate of 25%-100%, indicating its ability to balance efficiency and safety. At 50% Pene. Rate, CR achieves a higher average speed (14.18 m/s) compared to DR (12.06 m/s), but demonstrates a significantly smaller Min. Gap of 5.55 m versus DR's 49.04 m. This suggests CR prioritizes speed enhancement while tolerating reduced safety margins, whereas DR effectively suppresses risky overtaking behaviors through positional reward. Under 75% and 100% Pene. Rate, GR exhibits notably lower average speeds compared to normative levels, reflecting deficiencies in its reward design that induce vehicular stagnation phenomena, potentially due to excessive penalization of lane changes or conflicting intention resolution mechanisms.

Table 4: Metrics comparison of reward function in MADQN

| Pene. Rate | Rew. Fn. | Avg. Speed (m/s) | Min. Gap (m) | LC. Freq. (time/min) | Succ. Rate (%) |
|---|---|---|---|---|---|
| | GR | 4.38±1.20 | 1.85±1.64 | 0.34±0.06 | 30.15±2.49 |
| 25% | CR | **4.88±1.65** | 4.05±2.23 | **0.23±0.07** | 36.61±3.45 |
| | DR | 4.44±1.03 | **17.00±2.68** | 0.37±0.02 | **56.00±2.14** |
| | GR | 3.09±1.11 | 2.94±1.24 | 0.32±0.07 | 24.08±1.29 |
| 50% | CR | **5.57±2.64** | 6.50±2.37 | 0.34±0.10 | 38.14±3.63 |
| | DR | 4.66±2.13 | **28.99±0.75** | **0.28±0.03** | **43.75±1.16** |
| | GR | 5.67±3.33 | 6.42±3.79 | 0.38±0.12 | 35.86±1.04 |
| 75% | CR | 6.32±1.80 | 6.12±2.29 | 0.40±0.09 | 40.96±2.32 |
| | DR | **13.44±3.16** | **17.16±0.75** | **0.36±0.13** | **47.73±2.18** |
| | GR | 4.70±2.53 | 4.38±2.04 | **0.16±0.06** | 21.66±3.47 |
| 100% | CR | 7.61±3.50 | 7.10±4.64 | 0.45±0.15 | **46.07±2.87** |
| | DR | **9.89±2.48** | **15.10±0.72** | 0.34±0.03 | 39.96±0.18 |

Table 5: Metrics comparison of reward function in MAPPO

| Pene. Rate | Rew. Fn. | Avg. Speed (m/s) | Min. Gap (m) | LC. Freq. (time/min) | Succ. Rate (%) |
|---|---|---|---|---|---|
| | GR | 4.05±0.58 | 8.94±.69 | 0.42±0.07 | 30.97±4.55 |
| 25% | CR | 5.54±1.45 | 16.63±.14 | 0.29±0.03 | 28.32±4.13 |
| | DR | **8.18±3.19** | **14.56±0.49** | **0.03±0.02** | **72.43±0.15** |
| | GR | 4.06±1.71 | 10.35±.36 | 0.48±0.03 | 35.94±3.39 |
| 50% | CR | 5.63±3.99 | 16.28±.16 | **0.05±0.01** | 41.50±1.36 |
| | DR | **13.36±2.74** | **41.18±0.76** | 0.08±0.03 | **56.03±0.13** |
| | GR | 4.44±2.43 | 12.71±.19 | 0.21±0.06 | 23.42±3.67 |
| 75% | CR | 7.57±3.27 | 18.20±.43 | 0.14±0.02 | 41.11±1.25 |
| | DR | **14.27±2.36** | **49.51±0.75** | **0.11±0.03** | **60.06±0.16** |
| | GR | 4.80±2.82 | 13.11±.22 | 0.37±0.14 | 33.46±4.95 |
| 100% | CR | 6.43±4.03 | 27.65±.17 | 0.12±0.02 | 29.69±1.24 |
| | DR | **18.47±2.82** | **46.15±0.72** | **0.04±0.03** | **66.93±0.17** |

Regarding the LC. Freq., DR exhibits the lowest values except 100% Pene. Rate, demonstrating that the positional reward in Equation (23) effectively guides vehicles to stabilize in target lanes promptly, thereby reducing unnecessary lane changes. At 25%, 50%, and 75% penetration rates, GR and CR show significantly higher lane-changing frequencies than DR. This is hypothesized to result from smaller penalty weights on lane changes, where reward for lane-changing are frequently overshadowed by value function estimation errors, leading to inefficient maneuvers. Under the high penetration rate (100%) scenario, both CR and DR display notable increases in LC. Freq., which through analysis of test results is attributed to elevated CAV density, which induces reciprocal displacement among vehicles during interactive processes.

Tables 4 and 5 present the test results of MADQN and MAPPO respectively. The MADQN test data (Table 4) shows that, as indicated by the training curves, MADQN's overall performance is limited, with generally low Succ. Rate across all reward functions. This further confirms the inherent limitations of independent Q-learning-based MADQN in effectively handling multi-agent collaborative tasks. Nevertheless, the DR method demonstrates significant advantages in several key metrics. The most notable improvement lies in safety: under all penetration rates, the Min. Gap with DR policies is substantially greater than those of GR and CR. For instance, at a 25% penetration rate, DR achieves a minimum distance of 17.00m, compared to the dangerously small 1.85m and 4.05m for GR and CR respectively. This indicates that even when the base algorithm is unstable, DR's reward mechanism can effectively suppress risky behaviors. Additionally, DR consistently outperforms GR and CR in success rates, suggesting that DR can guide policy optimization toward more effective directions, despite the ultimate performance being constrained by the algorithm itself.

In Table 5, compared with MADQN, MAPPO demonstrates significantly improved overall performance, but the performance differences between reward functions become more pronounced. The DR function comprehensively and substantially outperforms GR and CR across all penetration rates and performance metrics. Particularly in terms of average speed and success rate, DR's advantages are most evident. For example, at 100% penetration rate, DR achieves an average speed of 18.47 m/s and a success rate of 66.93%.

An intriguing observation concerns the lane change frequency. Under the MAPPO algorithm, the lane change frequency with the DR strategy is extremely low, significantly lower than that of GR and CR (except at 100% penetration rate). This strongly demonstrates that the positional reward in DR effectively guides vehicles to stabilize quickly after reaching the target lane, reducing unnecessary or high-risk lane changes caused by inaccurate value function estimation. This result also explains why DR can maintain high traffic efficiency while achieving excellent minimum gap and success rates.

In summary, for different types of reinforcement learning algorithms, the reward differential method can ensure fundamental safety and improve task success rates. Under the QMIX algorithm, which exhibits superior overall performance, DR achieves success rates above 88% across penetration rates of 25%-100%, significantly outperforming CR and GR. The positional reward $r_p^i$ in DR is dynamically adjusted through gradient $\nabla f_p^i$, enabling vehicles to obtain higher rewards near target lanes. The algorithm involving the dot product of velocity and positional gradients, as shown in Equation (22), further guides vehicles to mitigate failures caused by trajectory deviations.

## VI. CONCLUSION AND FUTURE WORK

This paper proposes a differentiated reward method for reinforcement learning-based multi-vehicle cooperative decision-making algorithms. By incorporating gradient information of state transitions into reward function design, this approach addresses the limitations of conventional reward mechanisms in distinguishing action values within steady-state traffic flows. Experimental results demonstrate that the proposed method significantly enhances the convergence speed and training stability of multi-agent reinforcement learning algorithms. Strategies derived with the proposed reward function also outperforms those from centralized reward functions and general reward shaping methods across core metrics including traffic throughput, safety metrics, and action rationality. Furthermore, the differentiated reward method maintains stable performance across varying autonomous vehicle penetration rates, exhibiting robust scalability.

However, the differentiated reward method has limitations. Firstly, in value-iteration-type reinforcement learning algorithms, the gradient of the reward function amplifies action-value discrepancies, potentially inducing higher reward variance as evidenced in Figure 3 (such issues also exist in centering reward functions but manifest more severely in the reward-differential framework). This phenomenon may degrade overall algorithm performance. Additionally, our current implementation uses a discrete action space, which simplifies the gradient computation. Extending this to continuous control scenarios would require calculating higher-order vector derivatives of the reward function, posing a significant computational challenge.

Future work will focus on refining the theoretical foundations of differentiated reward function design to accommodate more reinforcement learning applications. Generalizability to More Complex Scenarios is also important. Our experiments were conducted on a multi-lane highway, the method's applicability to more complex traffic environments is still unknown. We plan to extend differentiated reward method to challenging scenarios such as unsignalized intersections, roundabouts, and merging zones. This will involve adapting the potential field's concept from a static "target lane" to a dynamic series of waypoints derived from trajectory planning, thus accommodating more fluid and conflicting vehicle objectives. Additionally, the current work demonstrates adaptability to varying CAV penetration rates but does not cover other critical environmental variables. A key priority will be to evaluate the method's performance under a wider spectrum of traffic conditions, from free-flow to congested traffic states, and under different traffic flow distributions. Furthermore, we will assess the algorithm's robustness against adversarial disturbances, such as sudden pedestrian intrusions or simulated sensor noise, to better evaluate its readiness for real-world deployment.